\documentclass[acmtog]{acmart}

\usepackage{booktabs} 
\usepackage{multirow}
\usepackage[ruled]{algorithm2e} 

\SetAlFnt{\small}
\SetAlCapFnt{\small}
\SetAlCapNameFnt{\small}
\SetAlCapHSkip{0pt}

\usepackage{color}
\newcommand{\pc}{\mathcal{P}} 
\newcommand{\cb}{\mathcal{C}} 
\newcommand{\vb}{\mathbf}
\newcommand{\kdr}[1]{\mathbb{R}^{#1}} 

\acmJournal{TOG}
\acmVolume{40}
\acmNumber{4}
\acmArticle{152}
\acmYear{2021}
\acmMonth{8}

\setcopyright{acmcopyright}

\acmDOI{10.1145/3450626.3459873}

\begin{document}
\title{Unsupervised Learning for Cuboid Shape Abstraction via Joint Segmentation from Point Clouds}

\author{Kaizhi Yang}
\affiliation{%
	\institution{University of Science and Technology of China}
	\city{Hefei}
	\country{China}
}
\email{ykz0923@mail.ustc.edu.cn}
\author{Xuejin Chen}
\affiliation{%
  \institution{University of Science and Technology of China}
  \city{Hefei}
  \country{China}
}
\email{xjchen99@ustc.edu.cn}

\begin{abstract}
Representing complex 3D objects as simple geometric primitives, known as shape abstraction, is important for geometric modeling, structural analysis, and shape synthesis. In this paper, we propose an unsupervised shape abstraction method to map a point cloud into a compact cuboid representation. We jointly predict cuboid allocation as part segmentation and cuboid shapes and enforce the consistency between the segmentation and shape abstraction for self-learning. For the cuboid abstraction task, we transform the input point cloud into a set of parametric cuboids using a variational auto-encoder network. The segmentation network allocates each point into a cuboid considering the point-cuboid affinity. 
Without manual annotations of parts in point clouds, we design four novel losses to jointly supervise the two branches in terms of geometric similarity and cuboid compactness.
We evaluate our method on multiple shape collections and demonstrate its superiority over existing shape abstraction methods. Moreover, based on our network architecture and learned representations, our approach supports various applications including structured shape generation, shape interpolation, and structural shape clustering.
\end{abstract}



\begin{teaserfigure}
  \centering
  \includegraphics[width=\linewidth]{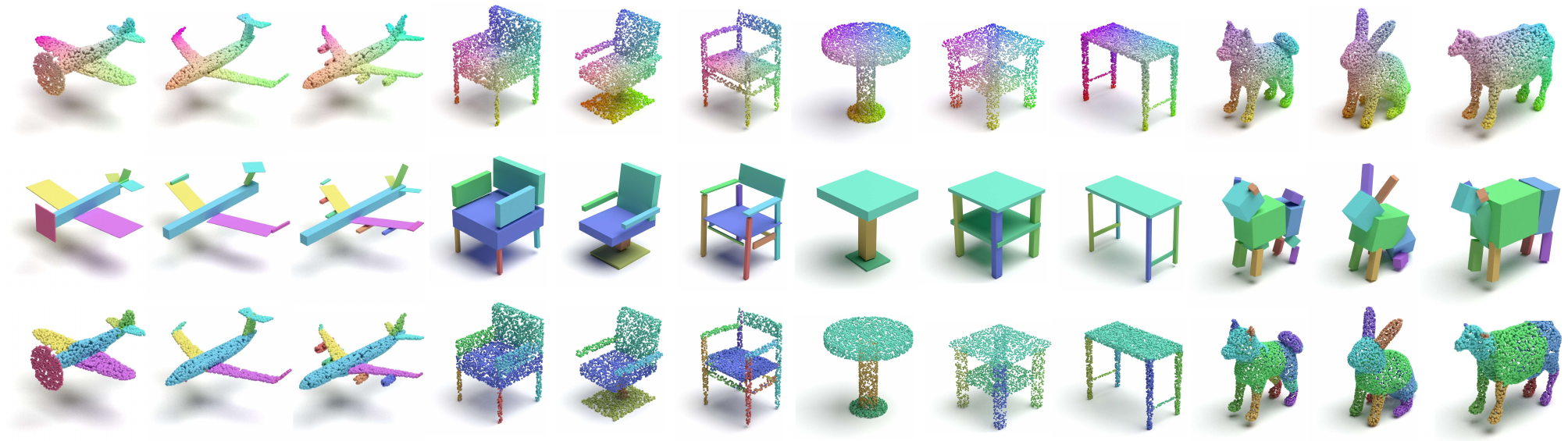}
  \caption{Our unsupervised approach supports cuboid-based shape abstraction and shape co-segmentation simultaneously. For different object categories, from a 3D point cloud (top row), our method predicts a set of parameterized cuboids (middle row) and segments the point clouds accordingly (bottom row). 
  The corresponding structural parts in different instances are presented in the same color. }
\end{teaserfigure}
\maketitle
\section{Introduction} \label{section:Introduction}
3D objects, especially manufactured objects, often have regularity in their external geometry. For example, chairs, typically contain regular parts such as the seat and back to achieve their functionality of stable
placement. These design rules can be reflected on the object structure, including a specific part shape or the relationship among parts.
Understanding objects from the structural perspective makes it easier to understand "why they are designed this way" or "why they look the way they do". While early vision tasks tend to understand an object from a holistic geometric perspective, some recent works
attempt to incorporate more structural information into 3D shape understanding or modeling tasks \cite{li_sig17,mo2019structurenet,Jie20DSMNet}.
Though these structure-based approaches have shown significant advantages, object structure, which is often hidden behind the geometry, is not easy to acquire. 
Many researchers have attempted to use manual annotations to attach structural information to the object shape \cite{Yi16, Mo_2019_CVPR}, due to the fact that humans can easily recognize the structure of objects. However, the automatic acquisition of structural representations for common objects is a challenging task.

In this paper, we focus on unsupervised shape abstraction, which dedicates to parsing objects into concise structural representation. Since the geometry of each part is much simpler than that
of the whole, shape abstraction attempts to use simple geometric primitives to assemble an object.
Previous works often employ data like volumetric \cite{abstractionTulsiani17}, watertight mesh \cite{sun2019abstraction}, or signed distance field \cite{smirnov2020dps} as the supervision information, since spatial occupancy information can provide important guidance for the shape abstraction task. However, these data are not intuitive to acquire and require extra geometric processing.

\begin{figure}[h]
  \centering
  \includegraphics[width=0.9\linewidth]{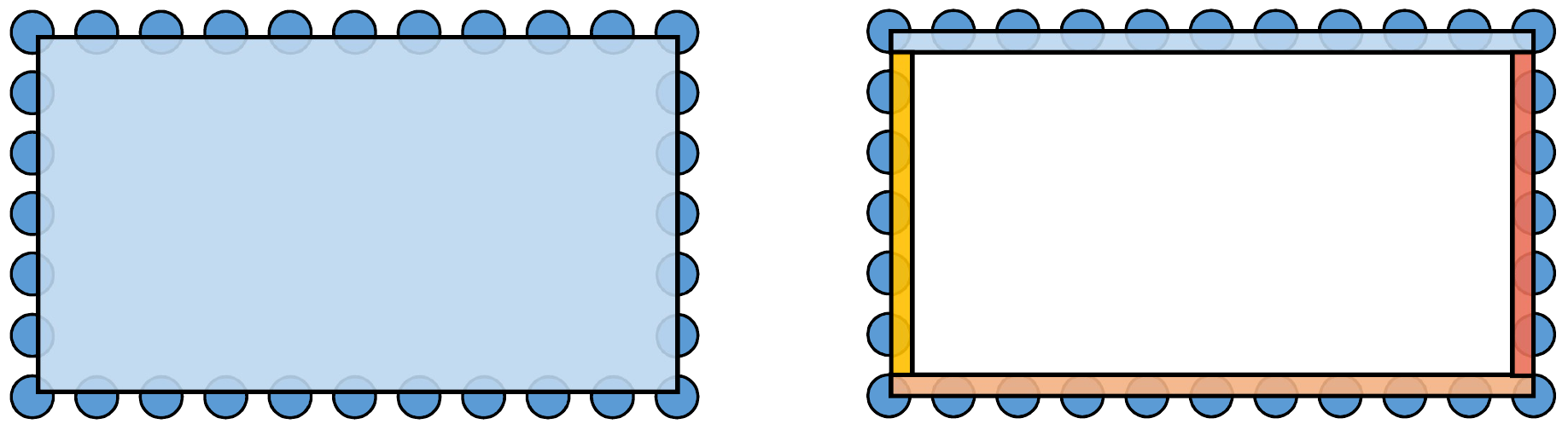}
  \caption{Ambiguity and degradation problems in the shape abstraction task for point clouds. The cuboid abstraction results on both sides well fit the point cloud in geometry. Compared to the left side which abstracts the point cloud as one single cuboid, the representation at the right side degenerates to surface fitting with a set of planes and ignores its structure.}
  \label{fig:Polysemy}
\end{figure}

On the other hand, point clouds are more similar to the raw scan data which is much easier to acquire by LiDAR or RGBD cameras.
However, little work has been done on learning primitive representations only through point clouds, since discrete point clouds lack dense occupancy supervision. The sparsity of point clouds easily causes degeneration in structural representation.
Fig.~\ref{fig:Polysemy} shows a 2D example. Considering a point cloud sampled from the surface of a rectangle, 
the two cuboid representations match its surface geometry as we can divide these points into arbitrary groups and fit them with different cuboids separately. Thus polysemy is introduced.
In this case, the shape abstraction task degenerates to surface fitting task with planes, and each primitive loses its structural properties.

Compared to shape abstraction which focuses more on geometry fitting, the co-segmentation task of point clouds is biased towards extracting the common structure of the whole dataset in order to prevent ambiguity and degeneration. 
Our main idea is to learn a general rule for assigning points to different cuboid primitives consistently among shapes, rather than directly based on the geometry proximity between the points and cuboids.
Moreover, for the joint tasks of segmentation and structural abstraction without structural annotations, we design several easy-to-implement loss functions that do not need any sampling operation on parametric surfaces. 
 
Under the unsupervised learning framework, we propose a shape abstraction network based on a variational autoencoder (VAE) to embed the input point cloud into a latent code and decode it into a set of parameterized cuboids. 
In addition, we make full use of the high-level feature embedding to exploit the point-cuboid correlation for the segmentation task.
Benefited from the joint segmentation branch, although some supervisory information is lacking, our algorithm is able to extract finer object structures than previous methods and obtain consistent part segmentation among instances with diverse shapes in the same category. 
Based on the latent space embedding, our method not only supports abstract shape reconstruction from point clouds but also supports shape interpolation and virtual shape synthesis by sampling the latent shape code.

In summary, this paper makes the following key contributions:
\begin{itemize}
\item We propose an unsupervised framework that entangles part segmentation and cuboid-based shape abstraction for point clouds.
\item A set of novel loss functions are designed to jointly supervise the two tasks without requiring manual annotations on parts or structural relationships.
\item In addition to shape abstraction and segmentation, our framework supports a range of applications, including shape generation, shape interpolation, and structural shape clustering.
\end{itemize}

\section{related work}
Manufactured objects always exhibit strong structural properties.
Understanding the high-order structural representation of shapes has been a hot research topic in the field of geometric analysis. 
The main difficulty of this task is that structural properties are embedded within the geometric shape.
Moreover, structure and geometry are intertwined and affect each other, making it challenging to decouple part structures through geometric shapes. 
In this section, we discuss the most relevant works on both supervised and unsupervised learning for object structures and shape co-segmentation. 

\paragraph{Supervised learning for object structure}
Some researchers have paid their attention to supervised learning for object structures from large-scale datasets with manual structural annotations \cite{Yi16, Mo_2019_CVPR} or decomposed parts using traditional optimization algorithms~\cite{zou20173d}.
These approaches can be divided into three categories according to the way how object parts are organized.
Approaches in a sequential manner \cite{zou20173d, Wu_2020_CVPR} employ recurrent neural networks to encode and decode parts sequentially.
In the parallel manner, parts are organized in a flat way \cite{G2L18, gaosdmnet2019, SAGnet19, Schor_2019_ICCV, dubrovina2019composite, li2020learning, gadelha2020learning}. 
Tree-based manner has attracted a lot of attention recently. Different levels in the tree structure represent different granularity of parts and parent-child node pairs represent the inclusion relationship between parts. 
Most tree-based methods require ground truth of the relationship between parts like adjacency and symmetry to construct the tree.
A generative recursive neural network (RvNN) is designed to generate shape structures and trained with multiple phases for encoding shapes, learning shape manifolds, and synthesizing geometric details respectively~\cite{li_sig17}.
Niu et al.~\shortcite{niu_cvpr18} employ an RvNN to decode a global image feature in a binary tree organization recursively.
A similar binary tree structure is also used for point cloud segmentation to utilize structural constraints across different levels \cite{yu2019partnet}.
In order to encode more elaborate structural relationships between parts, StructureNet is introduced to integrate part-level connectivity and inter-part relationships hierarchically in a $n$-ary graph~\cite{mo2019structurenet}.
Yang et al.~\shortcite{Jie20DSMNet} design a two-branch recursive neural network to represent geometry and structure in 3D shapes explicitly for shape synthesis in a controllable manner.

\paragraph{Unsupervised structural modeling}
On the other hand, many approaches use unsupervised learning for the structure parsing task assuming that the object geometry naturally reveals structural properties.
One direction is unsupervised shape abstraction, which ensembles 3D shapes by geometric primitives while preserving consistent structures in a shape collection.
Tulsiani et al.~\shortcite{abstractionTulsiani17} make the first attempt to apply neural networks for abstracting 3D objects with cuboids without part annotations. A coverage loss and a consistency loss are developed to encourage mutual inclusion of target objects and predicted shapes. 
Sun et al.~\shortcite{sun2019abstraction} propose an adaptive hierarchical cuboid representation for more compact and expressive shape abstraction. They construct multiple levels for cuboid generation in different granularities and use a cuboid selection module to obtain the optimal abstraction.
Besides cuboids, representing 3D shapes with more types of parametric primitives has been studied recently.
Smirnov et al.~\shortcite{smirnov2020dps} define a general Chamfer distance in an Eulerian version based on distance field, which allows abstraction of multiple parametric primitives.
Superquadric representation is another option for enhancing the geometric expressiveness and is demonstrated easier to learn than the cuboid representation for curved shapes~\cite{Paschalidou2019CVPR}.
They only use point clouds as supervision by constructing a bidirectional Chamfer distance between the predicted primitives and the point cloud.
However, it requires differentiable uniform sampling over the primitives, which is not easy.
In this paper, we design a new formulation of single-direction point-to-cuboid reconstruction loss that avoids sampling points on the cuboid surfaces.

\paragraph{Unsupervised segmentation}
Some unsupervised object segmentation works can also be used to analyze object structures.
Chen et al.~\shortcite{chen2019bae_net} treat co-segmentation as a shape representation learning problem.
They learn multiple implicit field branches for representing individual parts of an object and preserving consistency of the segmented parts over the entire dataset.
Aiming to learn the dense correspondence between 3D shapes in an unsupervised manner, a novel implicit function is proposed to measure the correspondence degree between points in different shapes and to obtain co-segmentation results~\cite{NEURIPS2020_335cd1b9}. 
Lin et al.~\shortcite{SEG-MAT-2020} develop an efficient approach based on the medial axis transform to identify junctions between parts of a 3D shape for segmentation.
However, this method does not consider the relationship among shapes like the common structure.
In comparison, we jointly learn shape abstraction and segmentation in an unsupervised manner while preserving the structural consistency over different instances on the segmentation task.

\begin{figure*}[t]
	\centering
	\includegraphics[width=\textwidth]{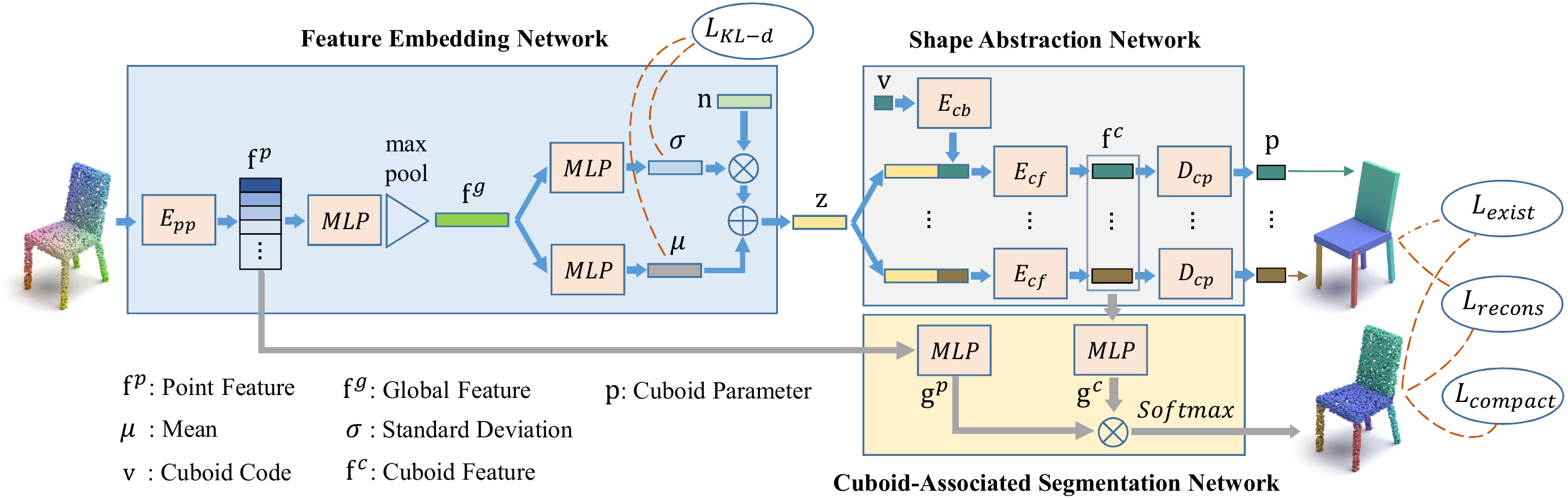}
	\caption{Overview of our joint shape abstraction and segmentation network. To connect the shape abstraction task and point segmentation task, our approach consists of three parts: the feature embedding network, shape abstraction network, and cuboid-associated segmentation network.
	As a result, our network obtains the cuboid-based structural representation and the point cloud segmentation results simultaneously.}
	\label{fig:Network}
\end{figure*}
\section{Our Approach}
\label{section:Network}
Given a point cloud $\pc$ that contains $N$ 3D points of an object, our goal is to reconstruct a set of cuboids $\{\cb_i\}_{i=1,\ldots,M}$ to concisely represent the 3D shape of the object. 
Similar to previous methods \cite{abstractionTulsiani17,sun2019abstraction}, each cuboid $\cb$ is parameterized into three vectors including a translation $\mathbf{t} \in \mathbb{R}^{3}$, a quaternion representing 3D rotation $\mathbf{r} \in \mathbb{R}^{4}$, and a scale $\mathbf{s} \in \mathbb{R}^{3}$. 

For different objects in various shapes of the same category, we attempt to predict a fixed number $M$ of cuboids. 
The fixed order of cuboids naturally conveys the part correspondence in different instances of the same category. 
However, even within the same category of objects, the structure of each instance varies a lot.
For example, some chairs have armrests while others do not. 
To fit different structures, we add an indicator for each cuboid $\delta \in \left\{ 0,1 \right\}$ to indicate whether this cuboid appears in an instance. 
In summary, we parameterize each cuboid $\cb_m$ as an $11$D vector $\vb{p}_{m} = [\mathbf{t}_{m};\mathbf{r}_{m};\mathbf{s}_{m};\delta_{m}]$ which includes the geometric properties and existence of the cuboid.

We employ a variational auto-encoder (VAE) \cite{DBLP:journals/corr/KingmaW13} framework that embeds the input point cloud into a latent code and decodes the parameters of the $M$ cuboids using a decoder, as shown in the feature embedding network and shape abstraction network in Fig.~\ref{fig:Network}. 
Meanwhile, we design a segmentation network that integrates an attention scheme to build the point-cuboid correlations for the point allocation for the $M$ cuboids.
To train the encoder-decoder network without manual annotations on part-based structures, we jointly learn the shape abstraction branch and the segmentation branch by enhancing the consistency between the part segmentation and reconstructed cuboids through a set of specially designed losses.

\subsection{Feature embedding network} 
The feature embedding network maps the input point cloud into a latent code $\vb{z}$ that follows a Gaussian distribution.
We first extract $N$ point-wise features $\vb{f}^{p}$ with $E_{pp}$ that contains two EdgeConv layers~\cite {DGCNN}, and each layer yields a $64$D feature for each point in each level.
By concatenating these two features together, the point feature $\vb{f}^{p}\in \kdr{128}$ is obtained. 
Then we extract a $1024$D global feature $\vb{f}^{g}$ by feeding the $N$ point features into a fully-connected layer with max-pooling~\cite{qi2016pointnet}.

As a generative model, we map the global feature $\vb{f}^{g}$ to a latent space. 
We design two branches composed of fully connected layers to predict the mean $\mu \in \mathbb{R}^{512}$ and standard deviation $\sigma \in \mathbb{R}^{512}$ of the Gaussian distribution of the latent variable, respectively. 
Then the latent code $\vb{z} \in \mathbb{R}^{512}$ is obtained by re-parameterizing a random noise $\vb{n} \in \mathbb{R}^{512}$ that follows to a standard normal distribution:
\begin{equation}{
    \vb{z} = \vb{\mu} + \vb{\sigma} \otimes \vb{n},
\label{equ:reparameterizing}
}\end{equation}
where $\otimes$ represents element-wise multiplication. 

\subsection{Shape abstraction network}
The shape abstraction sub-network decodes the latent code into the parameters for $M$ cuboids.
In order to retain the high-level structure information with part correspondence, we first infer $M$ cuboid-related features $\{\vb{f}^{c}_{m}\}_{m=1,\ldots, M}$ respectively by $M$ sub-branches for each cuboid from the latent code $\vb{z}$.
In the $m$-th branch, we embed a one-hot vector $\vb{v}_m$ where $\vb{v}_{mj}=1$ for $j=m$ and $\vb{v}_{mj}=0$ for $j\neq m $ with a cuboid code encoder $E_{cb}$ to obtain a $64$D embedding vector for the $m$-th cuboid. 
We concatenate the cuboid embedding vector with the latent code $\vb{z}$ and then fed into a cuboid feature encoder $E_{cf}$ to obtain a $128$D cuboid feature $\vb{f}^{c}_{m}$.
With the fixed order of cuboids and the one-hot cuboid codes, the part correspondences are preserved implicitly in the network so that the decoder not only contains information about the geometric shape information of the corresponding cuboids but also embeds the structure of a specific part in the object.

Then, each $\vb{f}^{c}_{m}$ passes through a cuboid parameter prediction module $D_{cp}$ to estimate the geometric parameters and existence probability for each cuboid.
Note that the feature encoders $E_{cb}$, $E_{cf}$, and the cuboid decoder $D_{cp}$ in each cuboid branch are all composed of fully connected layers and share parameters among all $M$ cuboids.

\subsection{Cuboid-associated segmentation network} \label{Co-segmentation-module}
 
The segmentation branch allocates each point in the input point cloud to $M$ cuboids. It means that we perform $M$-label point cloud segmentation, where each label corresponds to a cuboid, which is a potential part under the common structure of an object category.   
We use two fully-connected layers to reduce the dimension of point features $\vb{f}^p$ and the cuboid features $\vb{f}^{c}$ to $64$D feature vectors $\vb{g}^p$ and $\vb{g}^{c}$ respectively.
Treating these point features and cuboid features as the query and key in an attention scheme, we compute the affinity matrix $\vb{A}$ between $N$ points and $M$ cuboids as 
\begin{equation}
\vb{A}_{mn} = \vb{g}^{c}_m \cdot \vb{g}^{p}_n.
\end{equation}
Then a softmax operation is performed on each row to obtain the probability distribution that the point $p_{n}$ belongs to the part that the cuboid $C_m$ represents.
Thus, we get a probability matrix
\begin{equation}
   \mathbf{W}_{m,n} = \frac{ \exp{(\vb{g}^{c}_{m} \cdot {\vb{g}^{p}_{n}})} } {\mathop{\sum}\limits_{{m=1}}^{{M}}  \exp{(\vb{g}^{c}_{m} \cdot {\vb{g}^{p}_{n}})} }.
\label{equ:point-allocation-weight}
\end{equation}
From this probabilistic point-to-cuboid allocation matrix, we can simply obtain the segmentation result of a point cloud using argmax.

\subsection{Loss functions}
In order to train our network without ground-truth segmentation or cuboid shape representations, we design several novel losses between the results of the segmentation and abstraction branches to enforce the geometric coherence and structure compactness.
More specifically, we design a \emph{reconstruction loss} $L_{recons}$ between the segmentation and cuboid abstraction. 
A \emph{compactness loss} $L_{compact}$ is designed to encourage the network to learn a more compact structural representation based on the point-cuboid allocation in the segmentation branch.
The \emph{cuboid existence loss} $L_{exist}$ is designed to predict the existence indicator $\delta$ for each cuboid.
To enable the capability of shape generation, a \emph{KL-divergence loss} $L_{KL-div}$ is designed to enforce the latent code to follow a standard Gaussian distribution.

\subsubsection{Reconstruction loss} \label{section:Reconstruction_loss}
While no ground-truth cuboid annotations can be obtained in our unsupervised framework, the segmentation branch is utilized to provide a probabilistic part assignment for the input points. The reconstruction loss is expected not only to minimizing the local geometric distance but also encourages consistent high-order part allocations.
We calculate the distance $d(\vb{p}_n, \cb_m)$ between each point $\vb{p}_n$ of the input point cloud and each predicted cuboid $\cb_m$ and sum them with the probabilistic assignment predicted by the segmentation network as the shape reconstruction loss:
\begin{equation}
L_{recons}=\frac{1}{N}{\mathop{ \sum }\limits_{{n=1}}^{{N}}{{\mathop{ \sum }\limits_{{m=1}}^{{M}}{\mathbf{W}_{m,n}   d(\vb{p}_{n},\cb_{m}) }}}}.
\label{equ:reconstruction_loss}
\end{equation}

This loss function tends to reduce the geometric deviation for a point to the cuboid with high weights $\mathbf{W}_{m,n}$. 
In other words, this loss measures the compatibility of the segmentation branch with the abstraction branch. 
The shape parameters of a particular cuboid $\cb_{m}$ are optimized according to the weighted assignment $\mathbf{W}_{m,n}$, while the point-cuboid allocation probability $\mathbf{W}_{m,n}$ is adjusted according to the geometric shape of the cuboids.
Through this loss, we jointly optimize the cuboid parameters and the co-segmentation map.

Note that Eq. (\ref{equ:reconstruction_loss}) is actually a weighted single-direction distance from the point cloud to the cuboids.
Compared with the bidirectional Chamfer distance, it does not require a differentiable sampling operation on the cuboid surface.
Moreover, compared with training with Chamfer distance where points are assigned to the cuboids based on the distance between them, this formulation allows our model to jointly learn the cuboid assignment $\mathbf{W}_{m,n}$ explicitly, making it less likely stuck in a local minimum of geometric optimization.

\begin{figure}[h]
  \centering
  \includegraphics[width=0.9\linewidth]{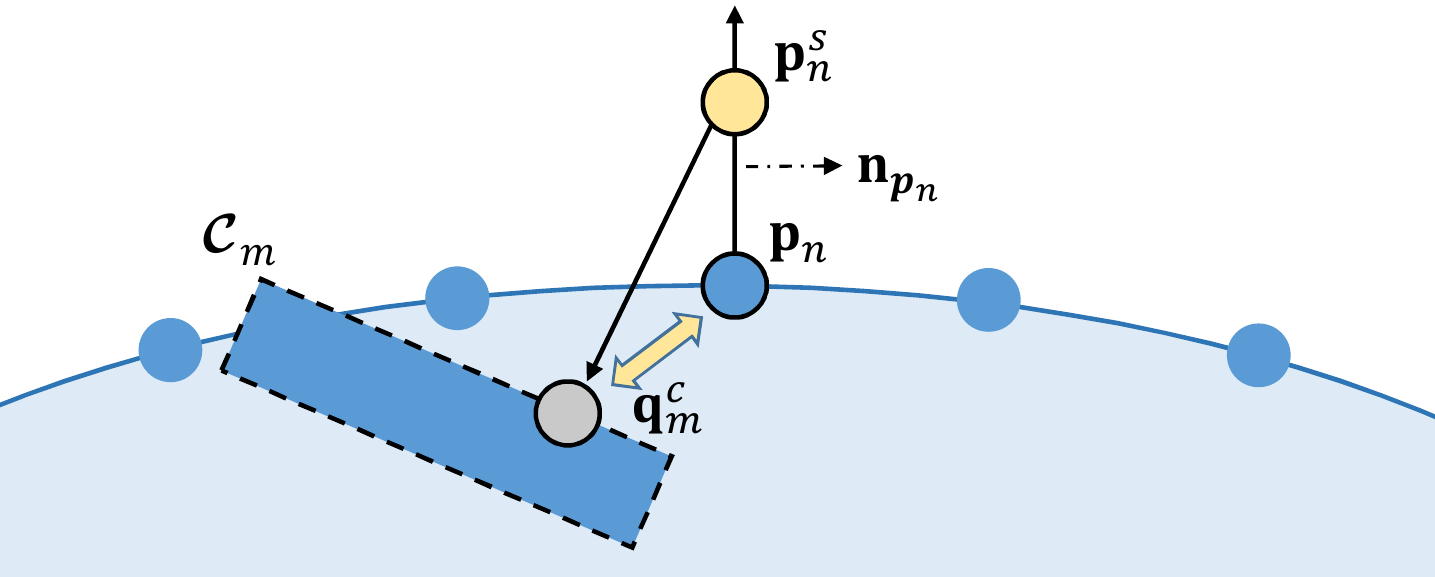}
  \caption{Illustration of our point-to-cuboid distance for calculating the reconstruction loss. For a point $\vb{p}_{n}$ in the input point cloud, we sample a point $\mathbf{p}^{s}_n$ along the normal direction $\vb{n}_{p_n}$ and look for its closest point $\vb{q}^{c}_m$ on the cuboid surface which has the most similar normal with $\vb{n}_{p_n}$. The Euclidean distance between $\vb{p}_{n}$ and $\vb{q}^{c}_m$ is defined as $d(\vb{p}_n,\vb{C}_m)$.}
  \label{fig:recon_loss}
\end{figure}

However, single-direction distances in general lead to model degeneration.
We introduce normal information for the reconstruction loss to prevent degradation.
Instead of calculating the distance from $\vb{p}_n$ to the closest cuboid plane, we calculate the distance from $\vb{p}_n$ to the cuboid surface with the most similar normal direction as $d(\vb{p}_{n},\cb_{m})$.
On the other hand, in order to emphasize the normal similarity and enhance the robustness to noises in point clouds, we introduce an additional sampling strategy when computing the distance.
We sample a new point $\mathbf{p}^{s}_n$ along the normal direction of $\vb{p}_n$ with a random distance from a Gaussian distribution $\mathcal{N}(0,\sigma_s^{2})$.
For the point $\mathbf{p}^{s}_n$, we look for its nearest point $\vb{q}^{c}_m$ on the selected cuboid surface and define the distance from point $\vb{p}_n$ to a cuboid $\cb_{m}$ as
\begin{equation} 
d(\vb{p}_n,\cb_{m}) = |\vb{q}^{c}_m - \vb{p}_n|^{2}.
\label{equ:reconstruction}
\end{equation}  
Fig.~\ref{fig:recon_loss} illustrates the distance definition considering both normal similarity and point sampling along the normal distribution. 
Under this definition, $d(\vb{p}_n,\cb_{m}) = 0$ only when $\vb{p}_{n}$ lies on the surface of the cuboid $\cb_{m}$ with same normal. Unless specifically mentioned, $\sigma_s$ is setting to 0.05 in our experiments.

\subsubsection{Cuboid compactness loss} \label{section:Sparsity_loss}
Typically, there exist multiple combinations of $M$ cuboids to represent an object. 
More cuboids tend to result in more accurate shape approximation.
However, the structure of an object is expected to be concise and clear.
Thus, a small number of cuboids is preferred for shape abstraction.
We design a cuboid compactness loss to penalize a large number of cuboids.

\begin{figure}[h]
  \centering
  \includegraphics[width=0.9\linewidth]{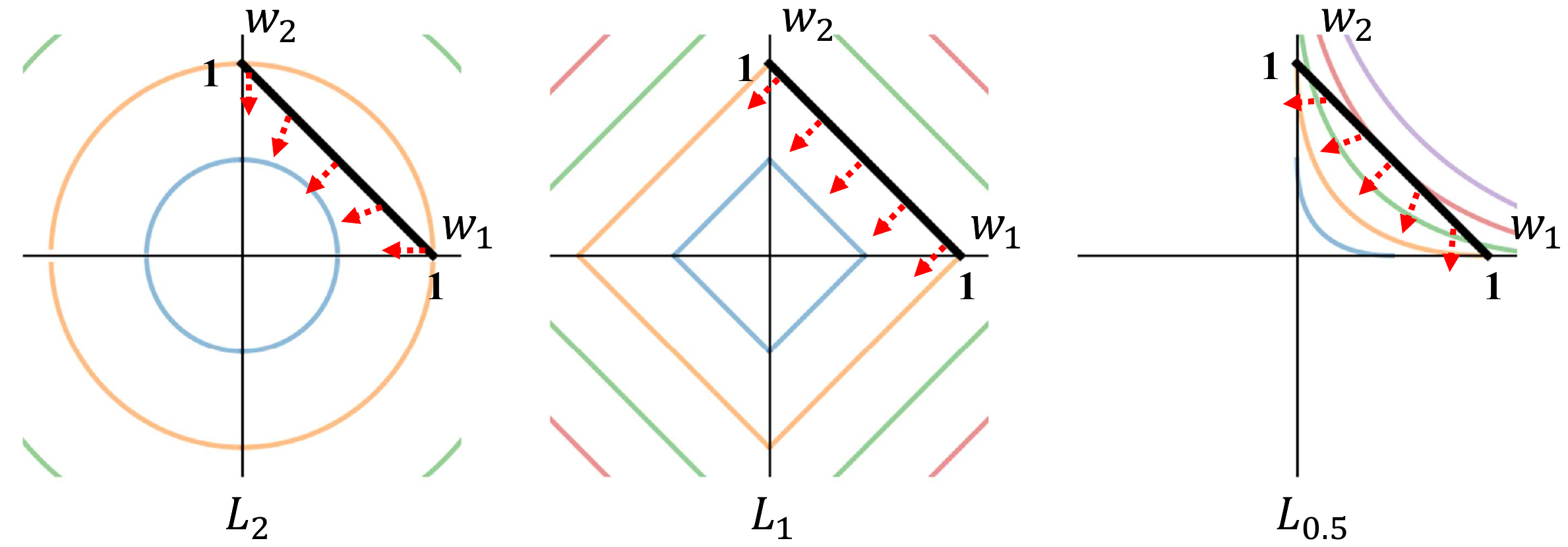}
  \caption{The illustration of $L_{2}$, $L_{1}$ and $L_{0.5}$ in the case of two cuboids, i.e. $w_1 + w_2 = 1$ (black line). The gradient of $L_{2}$ loss (red dotted line) will lead to $w_1 = w_2 = 0.5$ after optimization. The gradient of $L_{1}$ loss is vertical to the black line, making no update of $w_1$ and $w_2$ with the constraint $w_1 + w_2 = 1$. Minimizing $L_{0.5}$ leads to $w_1 = 0$ or $w_2 = 0$, i.e. the desired sparse solution.}
  \label{fig:L1}
\end{figure}

In the semantic segmentation task, when there is no label of a category in the segmentation result, one can consider that the object of that category does not appear.
Therefore, we impose a sparsity loss function on the segmentation result to reduce the number of cuboids used.
From our point-cuboid allocation probability matrix $\vb{W}$, we can compute the portion of how many points likely to be allocated to each cuboid as $w_m=\frac{1}{N} \sum_{{n=1}}^{{N}}{\mathbf{W}_{m,n}}$.
Analogously, if there are no points likely to be assigned to a certain cuboid $\cb_{m}$, i.e. $ w_m=0$, we regard the structure represented by the cuboid as absent in the 3D shape.
Therefore, we compute the compactness of the shape abstraction directly from $w_m$ for each cuboid.
Though $L_1$ loss is typically used to achieve sparseness, in our case $\sum^{M}_{m=1} w_m=1$ as defined in Eq.~(\ref{equ:point-allocation-weight}), the optimization process of $L_1$ loss does not update, as illustrated in Fig.~\ref{fig:L1}.
Instead, we adopt $L_{0.5}$ norm to compute the compactness loss as:
\begin{equation}
L_{compact} =  \Big( \sum_{{m=1}}^{{M}} \sqrt{w_m + \epsilon_{sps}} \Big) ^{2}.
\end{equation}
The small constant $\epsilon_{sps} = 0.01$ is added to prevent gradient explosion when a cuboid has no points allocated during training.

\subsubsection{Cuboid existence loss}
As mentioned in the parametric representation of the cuboids, we predict the existence of a certain cuboid so that we can allow a varied number of cuboids to represent the 3D shape for different instances. 
On the other hand, the point-cuboid segmentation branch can naturally handle a varying number of cuboids. 
Therefore, we consider the portion of points allocated to each cuboid $w_m$ after segmentation as its existence ground truth.
We set a threshold $\epsilon_{ext}$ so that when the number of points allocated to a cuboid $\cb_{m}$ by the allocation matrix  $w_m>\epsilon_{ext}$, we consider the cuboid existence $\delta^{gt}_{m} = 1$, otherwise $\delta^{gt}_{m} = 0$.
In our experiments, we set $\epsilon_{ext} = 0.05$.

We use binary cross-entropy between the predicted existence indicator $\delta_m$ for the cuboid in the shape abstraction sub-network and the approximated ground truth $\delta^{gt}_{m}$ as the cuboid existence loss:
 \begin{equation} 
    L_{exist} =  -\frac{1}{M} \sum_{m=1}^{M}{\big[\delta^{gt}_{m} \log \delta_{m} + (1-\delta^{gt}_{m}) \log (1-\delta_{m})\big]}
\end{equation}

\begin{figure*}[t]
  \centering
  \includegraphics[width=\textwidth]{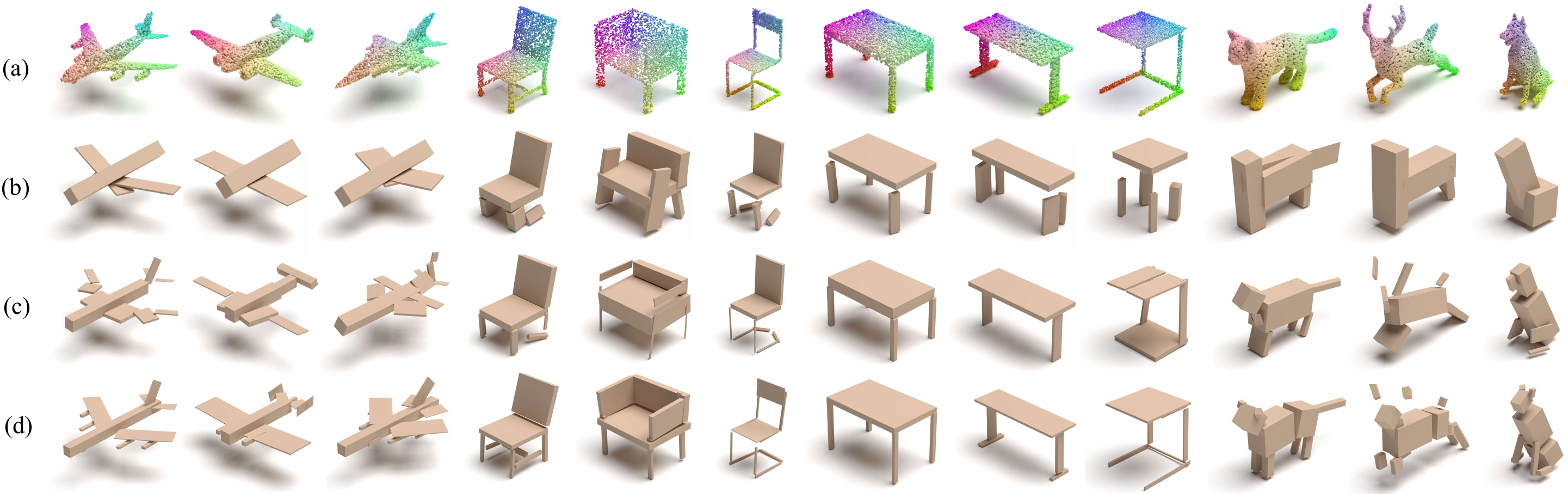}
  \caption{Comparison of cuboid reconstruction results. (a) Input point clouds. (b) Results of VP \cite{abstractionTulsiani17}. (c) Results of HA \cite{sun2019abstraction}. (d) Our results.
  VP tends to parse the target shapes into oversimplified structures.
  While HA uses a hierarchical representation, it is not easy to choose an appropriate grain for each shape, leading to over-partition or under-partition. 
  Both VP and HA fail to capture some small components, like the aircraft engines, brackets between the legs of chairs, etc. 
  In comparison, our method produces more realistic structures.}
  \label{fig:reconstruction}
\end{figure*}

\subsubsection{Latent code KL-divergence loss}
As in the vanilla VAE model \cite{DBLP:journals/corr/KingmaW13}, we also assume that the $512$D latent code $\vb{z}$ conforms to a standard Gaussian distribution with the assumption of independence of each dimension.
We use KL divergence as the loss function for distribution constraints:
 \begin{equation} 
    L_{KL-div} =  \frac{1}{2}\mathop{\sum}\limits_{{k=1}}^{{512}}{[e^{\sigma_{k}} - (1 + \sigma_{k}) + \mu_{k}^{2}]}.
\end{equation}
This VAE module with KL divergence loss supports shape generation and manipulation applications at a high-order structure level. 

\subsubsection{Network training}
We train our network end-to-end with the total loss
 \begin{equation}
        L = L_{recons} + \lambda_{1} \cdot L_{compact} + \lambda_{2} \cdot L_{exist} + \lambda_{3} \cdot L_{KL-div}
\end{equation}
We set $\lambda_{1} = 0.1$, $\lambda_{2} = 0.01$ and $\lambda_{3} = 6e-6$ in training. 
The network is implemented in the PyTorch framework \cite{NEURIPS2019_bdbca288}.
All the experiments were conducted on a NVIDIA Geforce GTX1080Ti GPU. 
We train each category separately.
The biases for the convolutional layers for predicting the cuboid scales and translations are all initialized to 0, and the one to predict rotation quaternions is initialized to $[1,0,0,0]$, while all the other parameters of the network are randomly initialized.
We use the Adam optimizer for a total of 1000 epochs, setting the batch size 32 and the learning rate $6e-4$.

\begin{figure*}[t]
  \centering
  \includegraphics[width=0.98\textwidth]{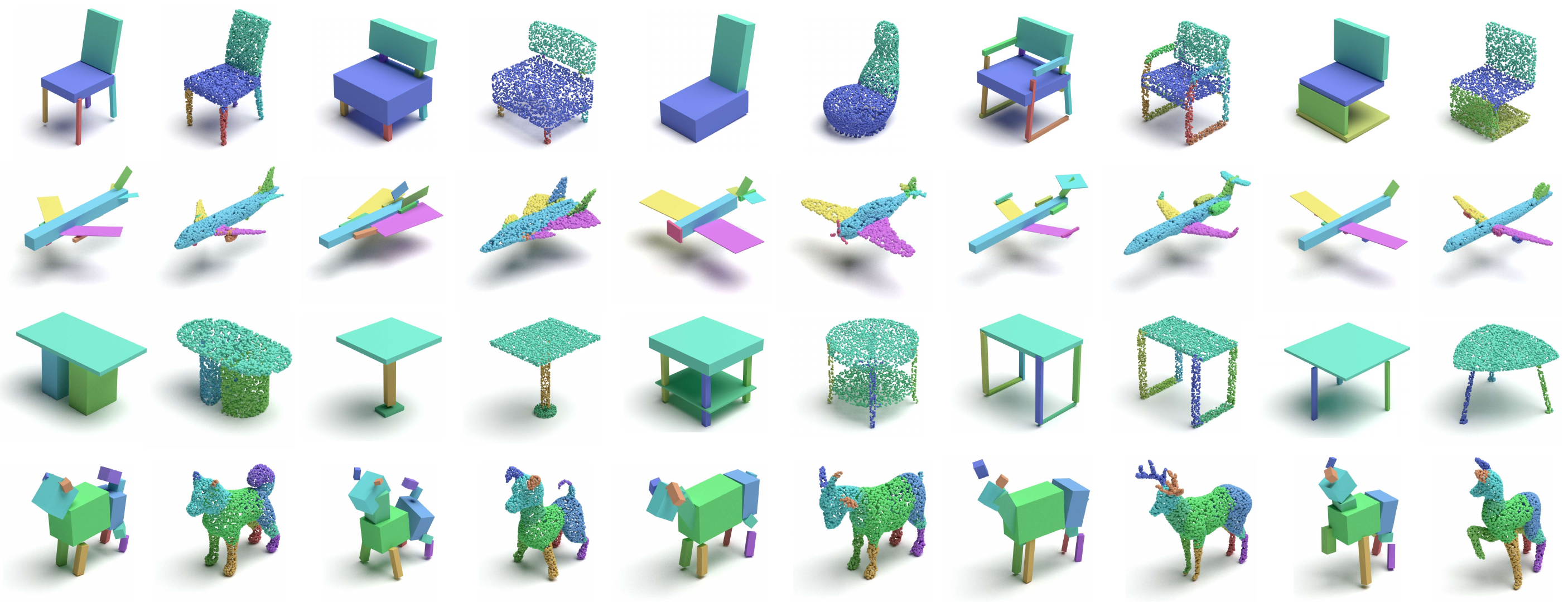}
  \caption{Visualization of our segmentation results on four object categories. The same color represents the cuboids of the same serial number. Although only simple cuboid representation is used, our method yields reasonable segmentation results for even non-cubic structures, such as aircraft fuselages, irregular seats, and animals. Our unsupervised segmentation results show consistency among shapes, such as seat backs, tabletops, etc., which are represented using the cuboids with the same order across examples.}
  \label{fig:segmentation}
\end{figure*}
\section{Experiments and Analysis}
Our joint shape abstraction and segmentation network obtains not only accurate structural representations but also highly consistent point cloud segmentation results. 
In this section, we mainly evaluate our method for the shape reconstruction task and the point cloud co-segmentation task and demonstrate its superiority compared to other shape abstraction methods.
Then we conduct ablation studies to verify the effectiveness of each component in our approach.

\subsection{Structured shape reconstruction}
\begin{table}[t]
	\caption{Quantitative comparison of shape reconstruction performance by Chamfer Distance. Our method outperforms the two state-of-the-art cuboid abstraction methods on the four categories.}
	\label{tb:ShapeRec}
	\begin{center}
		\begin{tabular}{l|cccc}
			\hline
    		Method & Airplane & Chair & Table & Animal \\
		    \hline
			VP~\cite{abstractionTulsiani17}  & 0.725 & 1.006 & 1.525 & 1.896 \\
			HA~\cite{sun2019abstraction}  & 0.713 & 1.109 & 1.449 & 0.575 \\
			Ours & \textbf{0.329} & \textbf{0.399} & \textbf{0.848} & \textbf{0.350} \\
			\hline
		\end{tabular}
	\end{center}
\end{table}

We first evaluate the shape reconstruction performance of our method and provide quantitative and qualitative comparisons to the previous cuboid-based shape abstraction methods.

For quantitative reconstruction evaluation, we use four categories of shapes: airplane (3640), chair (5929), table (7555) from ShapeNet dataset \cite{chang2015shapenet} and four-legged animal (129) from \cite{abstractionTulsiani17}. We divide the data into training data and test data in a 4:1 ratio same as \cite{sun2019abstraction}.
Due to the limited training data of the animal category, data augmentation with rotation by $\pm 5^{\circ}$ and $\pm 10^{\circ}$ around the $z$-axis for each model is performed.
For the four categories of shapes, we set $M$ to 20, 16, 14, and 16, respectively. All shapes are pre-aligned and normalized to the unit scale. 

We compare our method to two state-of-the-art cuboid-based shape abstraction methods: VP \cite{abstractionTulsiani17} and HA \cite{sun2019abstraction}. 
To evaluate the reconstruction performance, we adopt the commonly-used Chamfer Distance (CD) \cite{DBLP:conf/ijcai/BarrowTBW77} between two point sets.
The predicted point set is obtained by uniformly sampling over the predicted model composed of parametric cuboids. 
In our experiments, we use symmetric $L_2$ CD and evaluate on 4096 points sampled from the predicted model and the input point cloud.
Table \ref{tb:ShapeRec} shows the quantitative comparison of our method with other cuboid-based shape abstraction methods.
Our method outperforms the two state-of-the-art methods on the four object categories, demonstrating its capability for better geometric reconstruction by understanding various object structures. 
A group of reconstructed results using the three methods are shown in Fig.~\ref{fig:reconstruction}.
Both VP and HA methods have difficulty in capturing fine object structures, such as the armrests of chairs and the connection bars between the table legs.
In particular, VP tends to generate under-partitioned models.
On the other hand, due to the way of multi-level abstraction and selection in HA, some thin structures are forcibly divided into multiple small parts, such as chair legs.
In comparison, our method is able to extract more concise and precise results.

\subsection{Shape co-segmentation}
\begin{table}[t]
	\caption{Quantitative evaluation of shape co-segmentation performance. Our method performs the second-best while the best BSP-Net employs more complicated shape primitives. }
	\label{tb:ShapeSeg}
	\begin{center}
		\begin{tabular}{l|ccc}
			\hline
    		mIOU & Airplane & Chair & Table \\
		    \hline
			VP~\cite{abstractionTulsiani17}       & 37.6 & 64.7 & 62.1 \\
			SQ \cite{Paschalidou2019CVPR}     & 48.9 & 65.6 & 77.7 \\
			HA~\cite{sun2019abstraction}       & 55.6 & 80.4 & 67.4 \\
			BAE-Net \cite{chen2019bae_net} & 61.1 & 65.5 & 87.0 \\
			BSP-Net \cite{chen2020bspnet}  & \textbf{74.5} & \textbf{82.1} & \textbf{90.3} \\
			Ours     & 64.2 & 82.0 & 89.2 \\
			\hline
		\end{tabular}
	\end{center}
\end{table}
Though mainly designed for cuboid abstraction, our network also supports point cloud segmentation with part correspondences in one object category. 
In this section, we compare a number of shape decomposition networks on the task of unsupervised point cloud segmentation.
In addition to VP and HA approaches, we also compare with three state-of-the-art shape decomposition networks, Super Quadrics (SQ) \cite{Paschalidou2019CVPR}, Branched Auto Encoders (BAE) \cite{chen2019bae_net}, and BSP-Net (BSP) \cite{chen2020bspnet}.
SQ uses superquadrics as geometric primitives, which have more flexibility than the cuboid representation. 
BAE, on the other hand, uses the distance field as the base representation and generates a complex object jointly by predicting multiple relatively simple implicit fields. 
BSP is based on the binary space-partitioning algorithm that cuts the 3D space with multiple planes to depict the surface of a shape. The polyhedra composed of multiple planes can be used as primitives to represent object shapes. 
Note that a variety of training methods are introduced in BAE, while we choose to compare with its unsupervised baseline.

We conducted a quantitative comparison on three categories of shapes: airplane (2690), chair (3758), and table (5271) in the ShapeNet part dataset \cite{Yi16}.
Since we perform structural shape modeling, following \cite{chen2020bspnet}, we reduce the original semantic annotation in the dataset from (top, leg, support) to (top, leg) for the table category by merging the `support' label with `leg'. 
We adopt the mean per-label Intersection Over Union (mIOU) as the evaluation criterion for the segmentation task. 
Since the segmentation branch of our network does not refer to any semantics, we assign semantic annotations to each geometric primitive as following.
We first randomly take out 20\% of the shapes in the dataset, count the number of points belonging to a ground truth annotation for each primitive in the segmentation branch, and finally assign the ground truth annotation with the highest number of points as the label for that primitive. 
Afterward, these labels are transferred to the whole test set. 

The quantitative segmentation results on the co-segmentation task are compared in Table~\ref{tb:ShapeSeg}.
Our method achieves the best results within the cuboid-based approaches (VP, SQ, and HA) and ranks second among all the methods, demonstrating that our method generates segmentation results with higher semantic consistency.
Fig.~\ref{fig:segmentation} shows the segmentation results of our network. 
In addition to the above three categories, we also show segmentation results on the animal category.
Notice that our point cloud segmentation results are consistent with shape abstraction, which proves the effectiveness of our joint learning. 
Our method is able to subtly partition fine shape details, such as the engine of airplanes and the connection structure between the seat and the backrest.
Moreover, despite we adopt the cuboid representation, our network handles non-cubic structures well, for example, the fuselage of airplanes, the backrest and cushions of a sofa, and the animals.
In addition, it can be seen that our segmentation results exhibit a strong semantic consistency by using the same cuboid to express the same structure in different instances.

\subsection{Ablation study and analysis}
Our key idea is to learn the cuboid abstraction and shape segmentation jointly for mutual compensation.
To this end, we design a two-branch network with several loss functions to promote compatibility between them.
In this section, we disentangle our network and analyze the effect of each loss function by a group of ablation experiments.

\paragraph{Role of the segmentation module.} 
To train a shape abstraction model in an unsupervised manner, the most important thing is to assign parts of the input point cloud into corresponding primitives.
A reasonable part allocation will facilitate the learning process.
We explicitly learn the allocation weights $\vb{W}_{mn}$ in Eq.~\ref{equ:point-allocation-weight} by the segmentation branch.
To verify its effectiveness for unsupervised learning, we first remove the segmentation branch and directly assign points to its closest cuboid, i.e. $\vb{W}_{mn} = 1$ when $\cb_{m}$ is the closest cuboid of $\vb{p}_{n}$, otherwise $\vb{W}_{mn} = 0$. This variant is denoted as P2C-Dis.
Furthermore, we also train a model (Chamfer-Dis) without the segmentation branch but using the bidirectional Chamfer distance as reconstruction loss, as most previous unsupervised methods do.
For a fair comparison, we train the model with the segmentation branch using the reconstruction loss $L_{recons}$ only, denoted as P2C-Seg.

\begin{figure}[h]
  \centering
  \includegraphics[width=\linewidth]{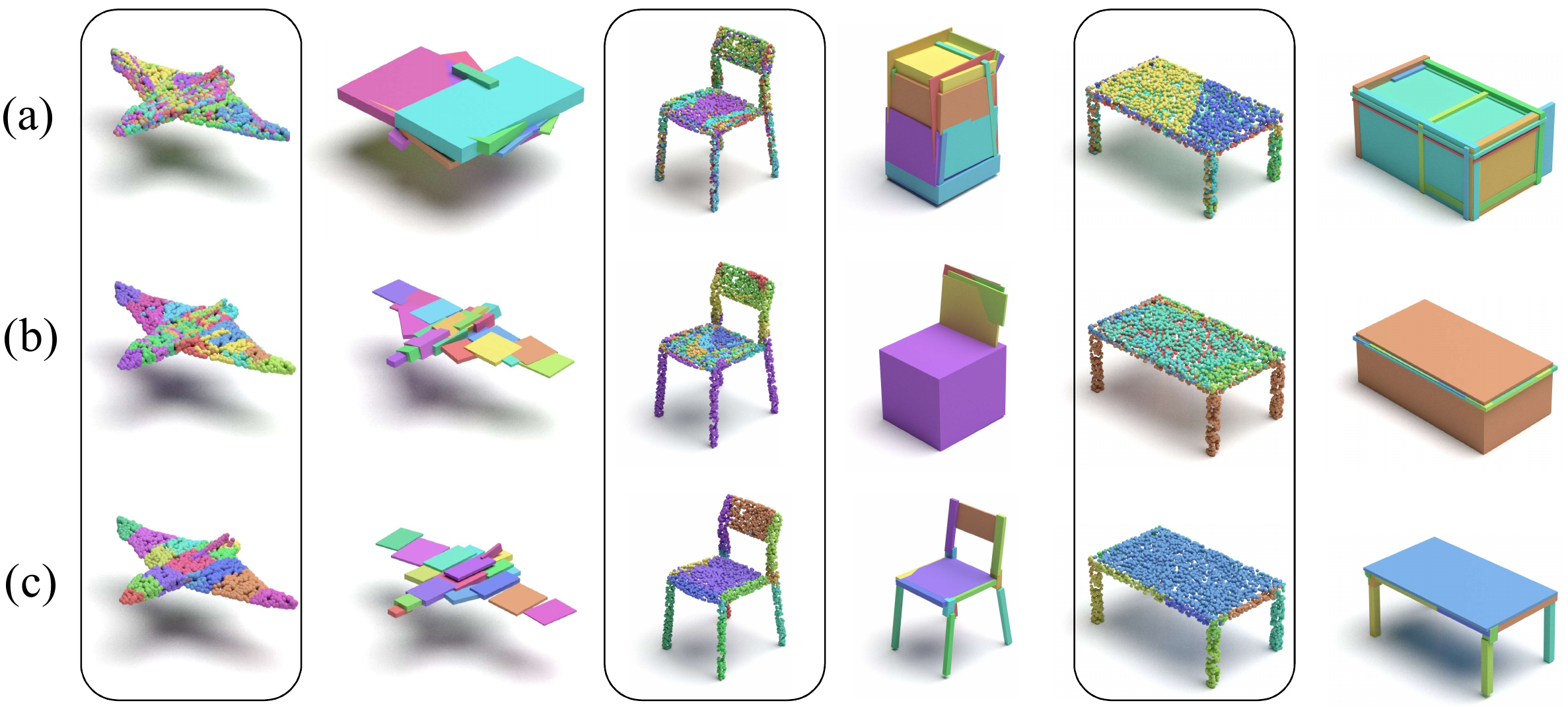}
  \caption{Cuboid assignment and abstraction results obtained from three variants with different reconstruction supervision. (a) P2C-Dis. (b) Chamfer-Dis. (c) P2C-Seg. The segmentation module in (c) helps shape abstraction with more reasonable part allocation.}
  \label{fig:Ablation_jointlearning}
\end{figure}

We compared the reconstruction results of the three variants in Table~\ref{tb:Ablation_jointlearning}, which shows that the P2C-Seg outperforms the other two variants on three categories.
In Fig.~\ref{fig:Ablation_jointlearning}, we visualize the part allocation results of the above three methods.
For P2C-Seg, we directly visualize the segmentation results predicted by the segmentation branch.
For Chamfer-Dis and P2C-Dis, as they assign a point to its closest cuboid primitive for computing reconstruction loss, we attach the label of the nearest cuboid to each point.
It shows that the part allocation results of P2C-Dis and Chamfer-Dis are more scattered than those of P2C-Seg, leading to stacked cuboids.
In addition, since it is difficult to adjust the hard assignment based on geometric distance, the P2C-Dis and Chamfer-Dis models usually get stuck in local minima.
For example, in the right column of Fig.~\ref{fig:Ablation_jointlearning}, the four chair legs are stuck into one large cuboid rather than four small cuboids.
In contrast, the part allocations of P2C-Seg are more explicit so that more compact abstraction results are achieved.

\begin{table}[t]
	\caption{Quantitative evaluation of cuboid reconstruction by Chamfer distance on three object categories using different supervision manners.}
	\label{tb:Ablation_jointlearning}
	\begin{center}
		\begin{tabular}{ccccc}
			\hline
    		 Category & P2C-Dis & Chamfer-Dis & P2C-Seg \\
		    \hline
			 Airplane & 4.836 & 0.286 & \textbf{0.237} \\
			 Chair & 3.356 & 0.834 & \textbf{0.436} \\ 
			 Table & 3.335 & 0.992 & \textbf{0.846} \\ 
			\hline
		\end{tabular}
	\end{center}
\end{table}

\paragraph{Point-to-cuboid distance.} 
\label{sec:Point2cuboid_distance}
The implementation of the point-to-cuboid distance $d(\vb{p}_{n},\cb_{m})$ in Eq.~(\ref{equ:reconstruction_loss}) greatly affects the network optimization process.
We make two designs on projection manner and random sampling to prevent model degeneration and enhance robustness to shape noises in $d(\vb{p}_{n},\cb_{m})$.
Next, we will disentangle these two designs and analyze their impacts respectively.

\begin{figure}[h]
  \centering
  \includegraphics[width=\linewidth]{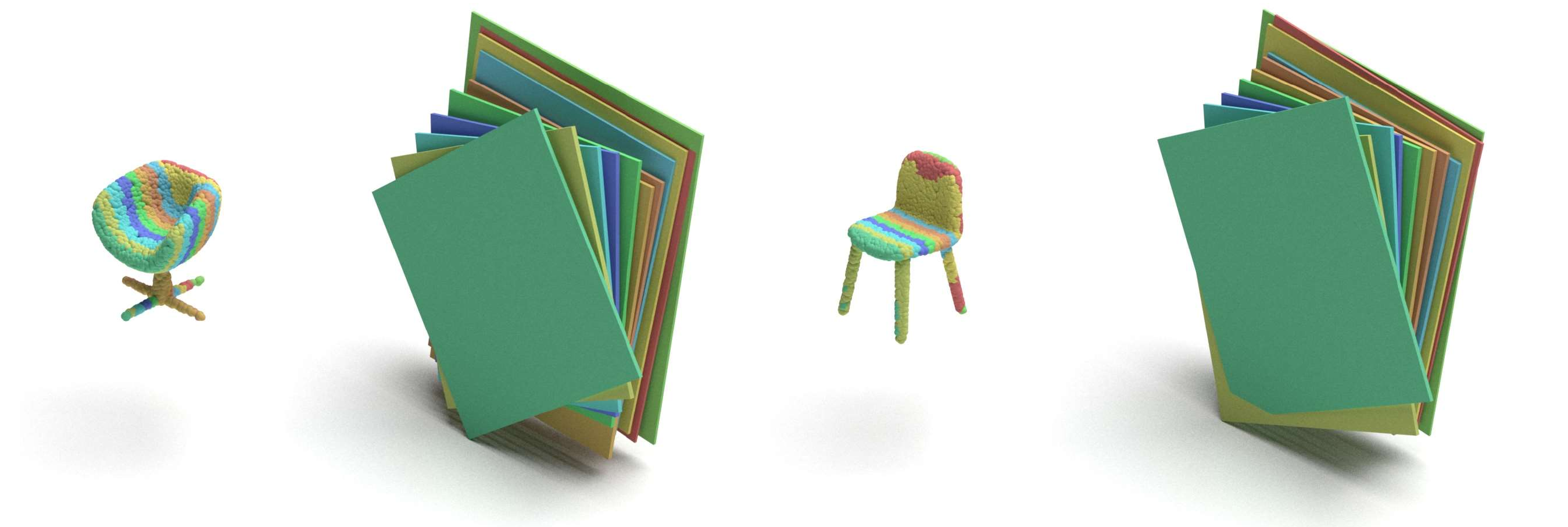}
  \caption{Degenerate solutions of minimum point-to-cuboid distance without considering normal consistency. The resulted cuboids tend to be thin and stacked in one direction.}
  \label{fig:Ablation_distance}
\end{figure}

An intuitive way of computing $d(\vb{p}_{n},\cb_{m})$ is to project a point $\vb{p}_{n}$ to its closest cuboid surface. 
However, as shown in Fig.~\ref{fig:Ablation_distance}, there exists a degenerate solution since the surface normal consistency is ignored.
This distance can be sufficiently small as long as there is at least one cuboid surface near each point. 
In contrast, the projection manner according to the normal direction not only ensures that the cuboid surface to be optimized has a similar surface normal with the point but also helps this normal-similar cuboid surface become the closest surface as the optimization proceeds.

While the projection manner selects which cuboid surface to compute the point-to-cuboid distance, we do not directly use the shortest distance from a point to a surface. 
Instead, we randomly sample a point $\vb{p}^s_n$, project it to the selected cuboid surface to find its closest point $\vb{q}^{c}_{m}$. 
The distance $|\vb{q}^{c}_m$-$\vb{p}_n|=0$ only when $\vb{p}_n$ lies on the surface of $\cb_{m}$ which has the same normal direction with $\vb{p}_n$. 
It enforces the network to emphasize geometric distance as well as normal similarity without increasing the computational complexity.
In contrast, the bidirectional Chamfer distance used in previous methods requires double calculation and sampling on cuboid surfaces. 
 
Moreover, this random sampling also improves the robustness to noisy point clouds. 
We add Gaussian noises to the input clouds with varying variance $\sigma_n^{2}$ and compare the reconstruction quality using different point-to-cuboid distances in Table~\ref{tb:Ablation_P2Cdis}.
The models trained with minimum-distance projection appear degenerate, while the models with normal-similar projection achieve satisfactory reconstruction accuracy.
The random sampling ($\sigma_s > 0$) drives the network to pay more attention to surface orientations, where the normal consistency is generally higher than those without point sampling along the normal direction ($\sigma_s = 0$).
In addition, the models trained with normal sampling show better robustness to noises. 
While $\sigma_n$ increases, the reconstruction quality using normal sampling during training is also generally better than those without sampling.
\begin{table}[t]
	\caption{Robustness of our point-to-cuboid distance. 
	Chamfer distance (smaller is better) and normal consistency (larger is better) between the reconstructed shape and the input point cloud under various noises $\sigma_n$ with different sampling range $\sigma_s$ are reported.}
	\label{tb:Ablation_P2Cdis}
	\begin{center}
		\begin{tabular}{c|c|ccc}
			\hline
    	    & Distance & \multicolumn{3}{c}{Normal}\\
    		\hline
			$\sigma_n$  & $\sigma_s = 0.000$ & $\sigma_s = 0.000$ & $\sigma_s = 0.025$ & $\sigma_s = 0.050$ \\
			\hline
			0.00  & 56.042/0.451 & \textbf{0.386}/0.758 & 0.389/0.762 & 0.399/\textbf{0.763} \\
			0.01  & 57.371/0.448 & 0.463/0.732 & 0.480/0.730 & \textbf{0.442}/\textbf{0.736} \\
			0.02  & 57.450/0.465 & 0.626/0.667 & 0.598/\textbf{0.679} & \textbf{0.572}/0.671 \\
			0.03  & 58.734/0.461 & 0.909/0.641 & \textbf{0.727}/0.645 & 0.754/\textbf{0.652} \\
			\hline
		\end{tabular}
	\end{center}
\end{table}

\begin{figure}[h]
  \centering
  \includegraphics[width=0.9\linewidth]{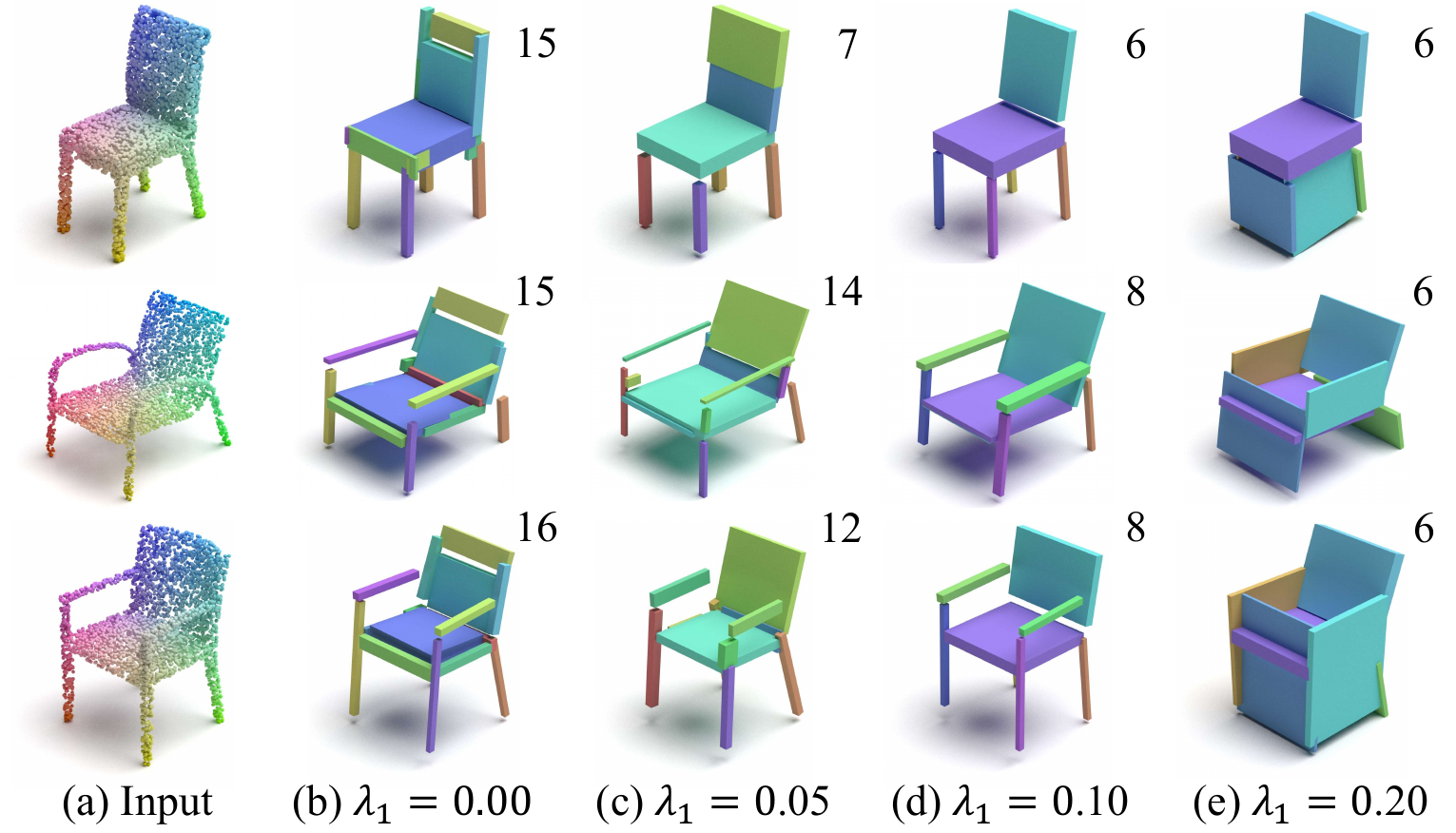}
  \caption{Abstraction results under different weights $\lambda_{1}$ for the compactness loss. We indicate the number of used cuboids on the right of each abstraction. Our method produces structurally consistent results in the same category with various weights. Suitable weights $\lambda_{1}$ lead to concise and accurately structured representations.}
  \label{fig:Ablation_compactness}
\end{figure}

\paragraph{Compactness loss.} 
The compactness loss is designed to penalize redundant cuboids.
We adjust the weight $\lambda_{1}$ exponentially for the compactness loss and analyze the reconstruction results.
In Table~\ref{tb:Compactness}, we report the average number of cuboids ($N_{AC}$) used in the shape abstraction results for all the instances, reconstruction quality (CD), and mIOU for part segmentation.
As $\lambda_{1}$ increases, $N_{AC}$ gradually decreases as analyzed in Sec. \ref{section:Sparsity_loss} and the CD increases due to the limited cuboids to represent the shape.
Fig.~\ref{fig:Ablation_compactness} shows the abstraction results of three shapes with different weights $\lambda_{1}$.
When $\lambda_{1}=0$ (b), the network is freely optimized without constraint on the number of cuboids, leading to over-disassembled shapes with redundant cuboids.
From (a) to (d), the increasing $\lambda_{1}$ leads to more concise structural representation.
When the $\lambda_{1}=0.20$, the network is too stingy with the use of cuboids resulting in the loss of some structure, which also is reflected in the CD and mIOU in Table~\ref{tb:Compactness}.
On the other side, the abstraction shapes are structurally consistent among multiple instances of the category under different settings of $\lambda_{1}$.
In each column, the common parts of different instances, such as the back, the seat, and four legs, are consistently represented.
\begin{table}[t]
	\caption{Abstraction performance with different weight $\lambda_{1}$ for the compactness loss.}
	\label{tb:Compactness}
	\begin{center}
		\begin{tabular}{ccccc}
			\hline
    		$\lambda_{1}$  & 0.00 & 0.05 & 0.10 & 0.20\\
		    \hline
			$N_{AC}$  & 15.486 & 11.144 & 9.753 & 5.544 \\
			CD  & 0.381 & 0.385 & 0.399  & 1.452 \\
			mIOU  & 81.2 & 82.1 & 82.0 & 72.5 \\
			\hline
		\end{tabular}
	\end{center}
\end{table}

\paragraph{Choice of the cuboid number $M$} 
In our experiments, we set the number of cuboids $M$ empirically for different categories.
To evaluate the sensitivity of our method to $M$, we change $M$ while fix all the other hyper-parameters to train our model.
In Table \ref{tb:Ablation_Cuboidnumber}, we report the average Chamfer distance and the average number of used cuboid in the abstraction results $N_{AC}$ under different $M$ for three categories.
As expected, the CD decreases as $M$ increases since more cuboids can be used to deal with diverse structures and better fit fine shapes.
Actually, decreasing $M$ with fixed weight $\lambda_{1}$ is analogous to increasing $\lambda_{1}$ with fixed $M$ when training our network. Similar abstraction results can be obtained with Fig.~\ref{fig:Ablation_compactness}.
\begin{table}[t]
	\caption{Impact of the maximum cuboid number $M$ on the shape abstraction task. We present the average CD and the $N_{AC}$ for comparison.}
	\label{tb:Ablation_Cuboidnumber}
	\begin{center}
		\begin{tabular}{ccccccc}
			\hline
    		Category &  & 8 & 12 & 16 & 20 & 24\\
		    \hline
			\multirow{2}{*}{Airplane} & CD & 0.849 & 0.484 & 0.394 & 0.329 & 0.285\\
			 ~                        & $N_{AC}$ & 4.171 & 5.663 & 9.801 & 11.801 & 13.720\\
			 \hline
			\multirow{2}{*}{Chair}    & CD & 1.381 & 0.501 & 0.399 & 0.385 & 0.366\\
			~                         & $N_{AC}$ & 5.976 & 7.150 & 9.753 & 10.130 & 12.158\\
			\hline
			\multirow{2}{*}{Table}    & CD & 1.170 & 1.062 & 0.826 & 0.776 & 0.657\\
			~                         & $N_{AC}$ & 5.636 & 7.328 & 8.524 & 9.747 & 12.573\\
			\hline
		\end{tabular}
	\end{center}
\end{table}

\begin{figure}[h]
  \centering
  \includegraphics[width=0.9\linewidth]{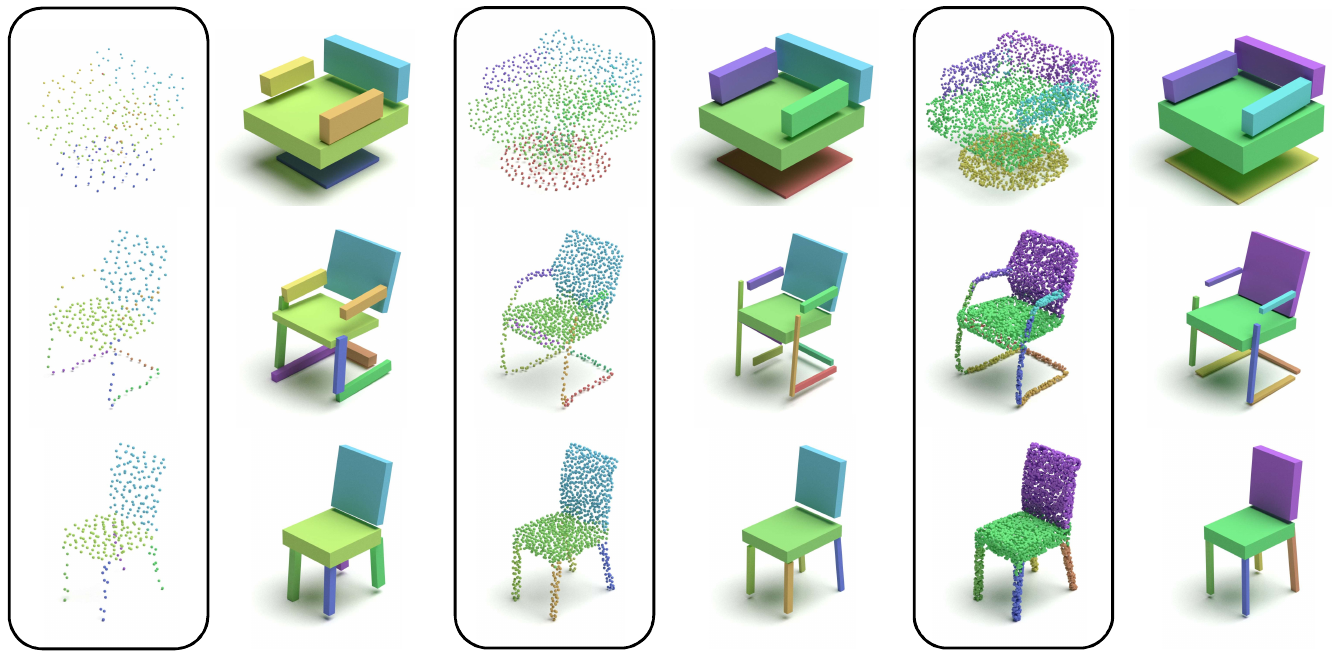}
  \caption{Segmentation and abstraction results trained with different point numbers. Reasonable structural representation with semantic consistency can be obtained even using fewer points.}
  \label{fig:Ablation_Sparsity}
\end{figure}

\paragraph{Robustness on sparse point clouds.} 
To verify the robustness of our framework to the density of input point clouds, we train our model with different numbers of points, including 256, 1024, and 4096, on the chair category. 
In Fig.~\ref{fig:Ablation_Sparsity}, we show the abstraction and segmentation results under different point cloud densities for the same shape.
It shows that our method produces reasonable and consistent structural representation for the same shape with various point densities.

\subsection{Applications}
Based on our network architecture, our method supports multiple applications of generating and interpolating cuboid representation, as well as structural clustering.
\begin{figure}[h]
  \centering
  \includegraphics[width=\linewidth]{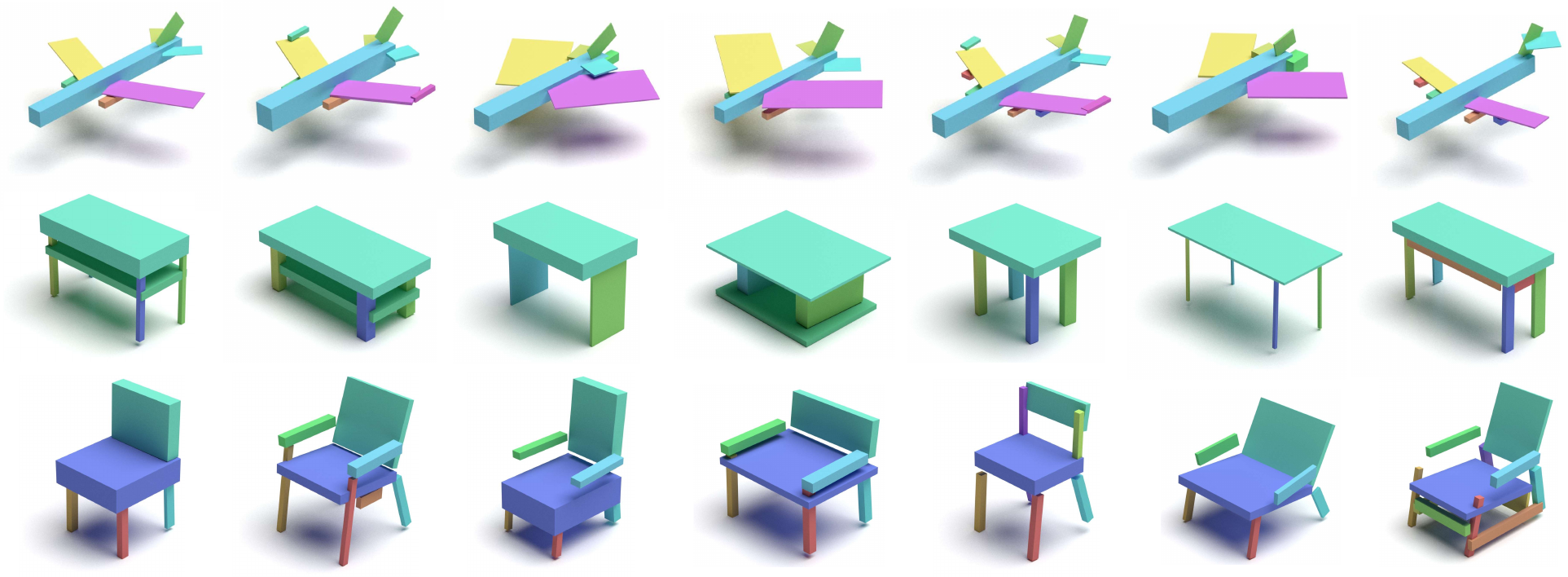}
  \caption{Shape generation results by randomly sampling a standard normal noise vector. The automatically generated shapes have reasonable and realistic cuboid structures with a large diversity.}
  \label{fig:generation}
\end{figure}
\paragraph{Shape generation}
Benefited from the VAE architecture, our network can accomplish the shape generation task by sampling the latent code $\vb{z}$ from a standard normal distribution, which is neglected in previous unsupervised shape abstraction methods, such as VP and HA.
Fig.\ref{fig:generation} shows a group of generated shapes, demonstrating the capability of our method in generating structurally diverse and plausible 3D shapes with cuboid representation.
%

\begin{figure}[h]
  \centering
  \includegraphics[width=\linewidth]{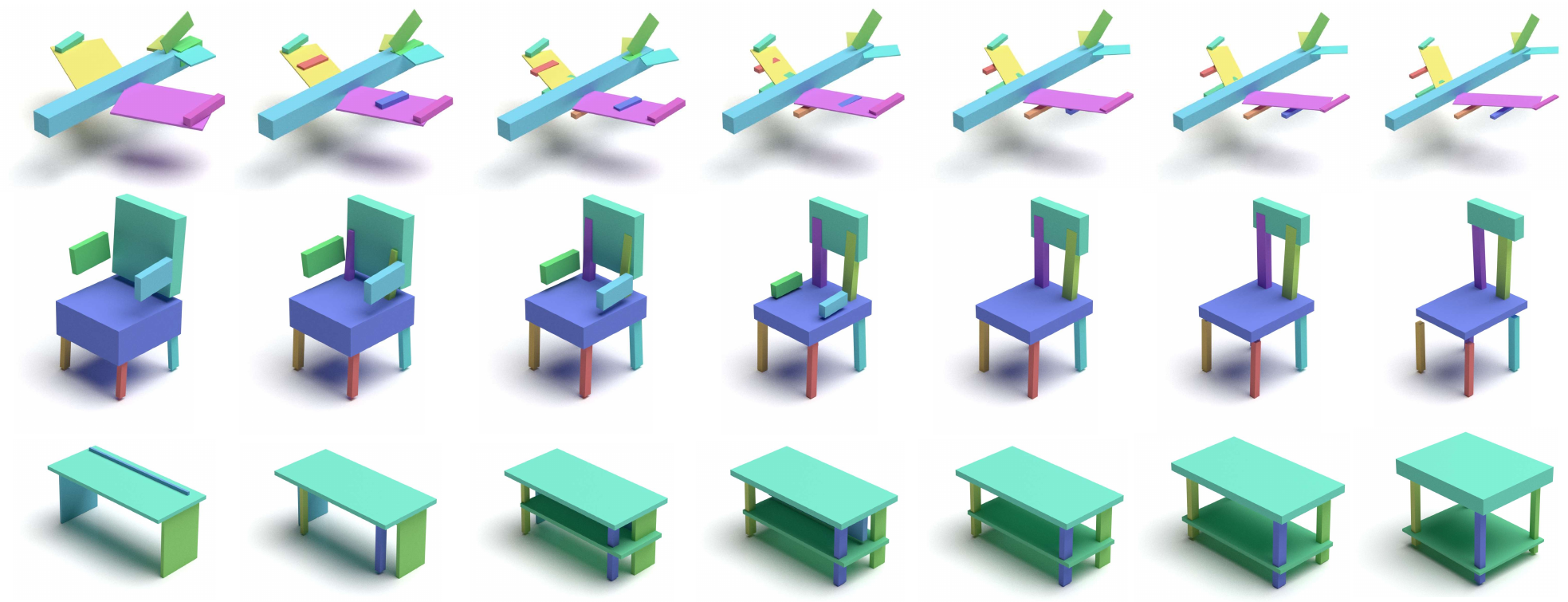}
  \caption{Shape interpolation by linearly interpolating between the latent codes $\vb{z}$ of two shapes. Continuous change in geometry and structure can be observed in the interpolated shapes from left to right.}
  \label{fig:interpolation}
\end{figure}

\paragraph{Shape interpolation} Shape interpolation can be used to generate a smooth transition between two existing shapes. The effect of shape interpolation depends on whether the network can learn a smooth high-dimensional manifold in the latent space. 
We evaluate our network by interpolating between pairs of 3D shapes to demonstrate that our method learns a smooth manifold for interpolation.
As illustrated in Fig.\ref{fig:interpolation}, our shape abstraction network produces a smooth transition between two shapes with reasonable and realistic 3D structures.
For example, the backrest gradually becomes smaller and the seat gets progressively thinner from left to right.

\paragraph{Structural shape clustering}
\begin{figure}[h]
  \centering
  \includegraphics[width=\linewidth]{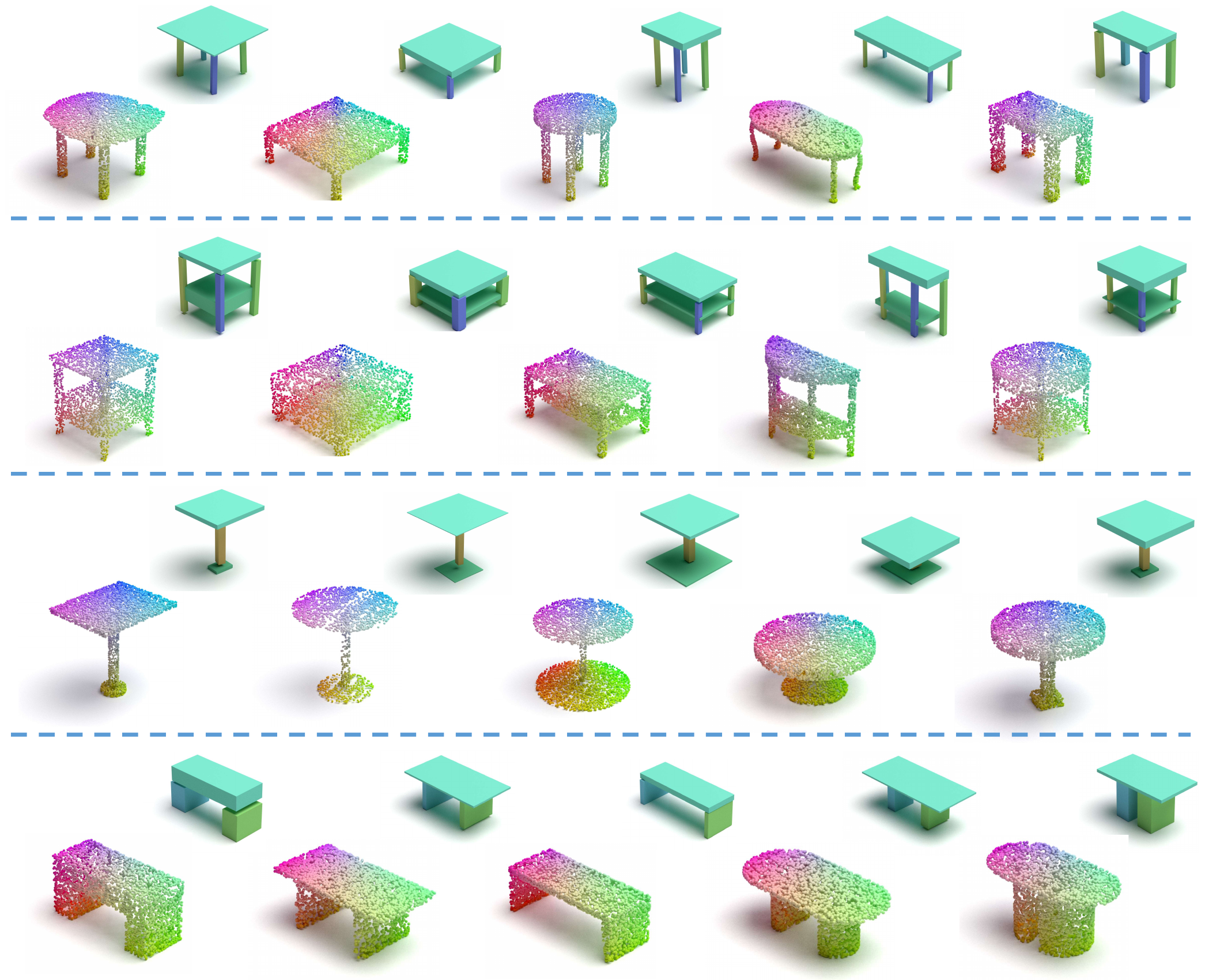}
  \caption{Four groups of shapes of the table category using our structural clustering. We use the learned cuboid existence indicators as measurement and group shapes that have the same existence indicator together. From top to bottom, the serial numbers of appearing cuboids are [5,7,10,12,14], [5,7,9,10,12,14], [3,9,10], and [7,10,12] respectively.} 
  \label{fig:classification}
\end{figure}
Though the object geometries can vary dramatically within the same category, they often follow some specific structural design. 
Based on the learned structured cuboid representation, our method supports 3D shape clustering according to the common shape structure. 
We use the existence indicator vector of the $M$ cuboids learned by our abstraction network to represent the object structure.
The shapes in which the same cuboids appear are considered to have the same structure. 
Fig.~\ref{fig:classification} shows four groups of structural clustering results of the table category. 
We can see that in each cluster, even though the geometric details vary greatly, the 3D shapes share a common structure.

\subsection{Failure cases, limitation and future works}
\begin{figure}[h]
  \centering
  \includegraphics[width=\linewidth]{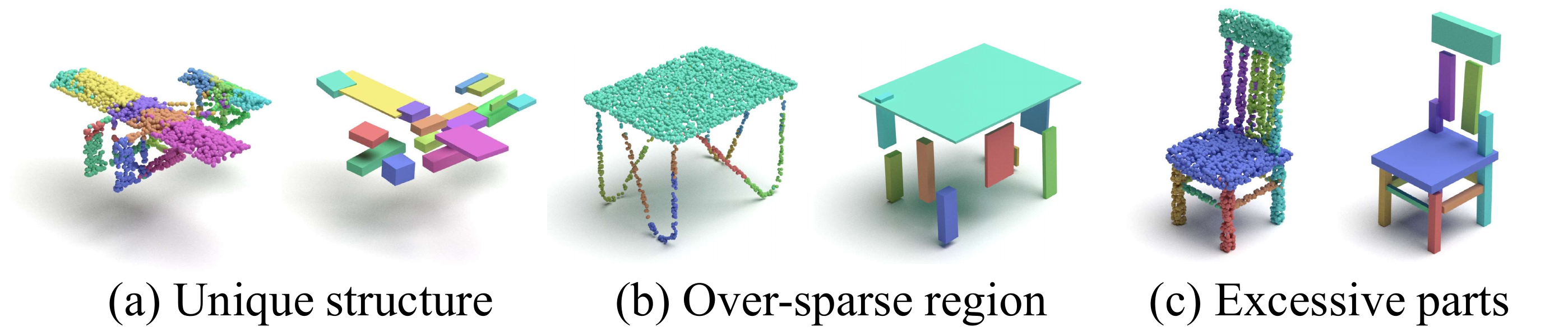}
  \caption{Three representative failure cases of our method.}
  \label{fig:Ablation_failure}
\end{figure}

The robustness and effectiveness of our unsupervised shape abstraction method have been demonstrated by extensive experiments. 
It also has some limitations and fails in some special cases.
Since our method takes the structure consistency between different shape instances in the same category into account, the instance with very unique structures may not be well reconstructed, such as the aircraft in Fig.~\ref{fig:Ablation_failure} (a). 
Due to the uniform sampling of point clouds in 3D space, points sampled for fine structures are too scarce to provide sufficient geometric information, e.g. the table legs in Fig.~\ref{fig:Ablation_failure} (b).
Another failure case is caused by a small maximum number of cuboid $M$.
For shapes with excessive small parts, our method fails to precisely recover all the parts with limited $M$. 
For the example in Fig.~\ref{fig:Ablation_failure} (c), the six thin slats are grouped into two parts in order to keep semantic consistency with other chairs.

Our framework can be improved in several directions in the future.
Currently, our shape abstraction network is trained separately for each specific object category. A future direction is to encode multiple classes of shapes in one network simultaneously or to learn transferable features across categories.
Another direction is integrating multiple geometric primitives to represent objects, or the primitives with stronger representational capability, such as polyhedra \cite{deng2020cvxnet} and star-domain \cite{kawana2020neural}.
However, there should be a trade-off between representational capability and structural simplicity.
Moreover, although the manually annotated dataset \cite{Mo_2019_CVPR} already contains rich relationships between parts, unsupervised relationship discovery among parts is still a challenging task, which can be further investigated.
\section{Conclusions}
In this paper, we introduce a method for unsupervised learning of cuboid-based representation for shape abstraction from point clouds.
We take full advantage of shape co-segmentation to extract the common structure in an object category to mitigate the multiplicity and ambiguity of sparse point clouds. We demonstrate the superiority of our method on preserving structural and semantic consistency in cuboid shape abstraction and point cloud segmentation. 
Our generative network is also versatile in shape generation, point cloud segmentation, shape interpolation, and structural classification.

\begin{acks}
We thank the anonymous reviewers for their constructive comments. We are grateful to Dr. Xin Tong for his insightful suggestions on this project. This work was supported in part by the National Natural Science Foundation of China (NSFC) under Grant 61632006 and Grant 62076230; in part by the Fundamental Research Funds for the Central Universities under Grant WK3490000003; and in part by Microsoft Research Asia.
\end{acks}

\bibliographystyle{ACM-Reference-Format}
\bibliography{main}

\end{document}